\documentclass[sigconf,natbib=true]{acmart}
\usepackage{pifont, multirow, multicol, graphicx, booktabs, subcaption, stfloats}
\AtBeginDocument{%
  }

\setcopyright{acmlicensed}
\copyrightyear{2026}
\acmYear{2026}
\acmDOI{XXXXXXX.XXXXXXX}
\acmConference[SIGIR '26]{the 49th International ACM SIGIR Conference on Research and Development in Information Retrieval}{July 20--24,
  2026}{Melbourne, Australia}
\acmISBN{978-1-4503-XXXX-X/2026/07}

\begin{document}

\newcommand{\ours}{{TEC}}

\title{\ours{}: A Collection of Human Trial-and-error Trajectories for Problem Solving}

\author{Xinkai Zhang}
\affiliation{
  \institution{College of AI, Tsinghua University}
  \institution{Quancheng Laboratory}
  \institution{Gaoling School of Artificial Intelligence, Renmin University of China}
  \city{Beijing 100084}
  \country{China}}
\email{stevenzhangx@ruc.edu.cn}

\author{Jingtao Zhan}
\authornote{Co-corresponding authors.}
\affiliation{%
  \institution{Institute of Data and Information,
Tsinghua Shenzhen International Graduate School}
  \city{Shenzhen 518055}
  \country{China}}
\email{jingtaozhan@tsinghua.edu.cn}

\author{Yiqun Liu}
\affiliation{%
  \institution{Department of Computer Science and Technology, Tsinghua University}
  \city{Beijing 100084}
  \country{China}}
\email{yiqunliu@tsinghua.edu.cn}

\author{Qingyao Ai}
\authornotemark[1]
\affiliation{%
\institution{Quancheng Laboratory}
  \institution{Department of Computer Science and Technology, Tsinghua University}
  \city{Beijing 100084}
  \country{China}}
\email{aiqy@tsinghua.edu.cn}

\renewcommand{\shortauthors}{Zhang et al.}



\begin{abstract}

Trial-and-error is a fundamental strategy for humans to solve complex problems and a necessary capability for Artificial Intelligence~(AI) systems operating in real-world environments. Although several trial-and-error AI techniques have recently been proposed, most of them rely on simple heuristics designed by researchers and achieve limited performance gains. The core issue is the absence of appropriate data: current models cannot learn from detailed records of how humans actually conduct trial-and-error in practice.
To address this gap, we introduce a data annotation platform and a corresponding dataset, termed \textbf{T}rial-and-\textbf{E}rror \textbf{C}ollection~(\ours{}). The platform records users’ complete trajectories across multiple trials and collects their reflections after receiving error feedback. Using this platform, we record the problem-solving processes of 46 participants on 58 tasks, resulting in 5,370 trial trajectories along with error reflections across 41,229 webpages.
With this dataset, we observe that humans achieve substantially higher accuracy compared to LLMs, which demonstrates that humans are more effective in trial-and-error than LLMs. We believe that the \ours{} platform and dataset provide a valuable foundation for understanding human trial-and-error behavior and for developing more capable AI systems. Platform and dataset are publicly available.\footnote{\url{https://github.com/Serendipity0429/TEC}}


\end{abstract}


\maketitle

\section{Introduction}a
Trial-and-error is a fundamental mechanism of natural selection: organisms that relentlessly try and learn from errors may survive; those that do not become extinct~\cite{popper1999all, darwin1859origin}. 
This iterative process demands two dimensions of intelligence.
The \emph{trial} aspect requires problem understanding, strategy selection, and effective tool use to explore candidate solutions~\cite{newell2019human,Simon1978InformationProcessingTO}. The \emph{error} aspect requires self-evaluation and reflection, recognizing \emph{why} a trial failed and deciding \emph{what} to change next~\cite{metcalfe2017learning,10.3758/s13423-021-02022-8}. 
This principle is equally important for AI research. When AI systems are expected to perform complex tasks on behalf of humans, these tasks are rarely solvable in a single trial. Instead, they often require iterative trials and the ability to learn from errors. Recent studies have begun to incorporate the idea of trial-and-error into LLM-based agents, such as inference scaling~\cite{Guo_2025,openai2024openaio1card,yang2025qwen3technicalreport,kimiteam2025kimik15scalingreinforcement} and reflection~\cite{zhan2025evaluatingintelligencetrialerror,yuan2025agentrtraininglanguagemodel,ma-etal-2025-s2r,cheng2025agentr1trainingpowerfulllm,gou2024criticlargelanguagemodels}, and have demonstrated substantial improvements in model capability.

However, there is a notable lack of data to demonstrate human trial-and-error processes. As a result, current AI approaches largely rely on relatively simple heuristic rules designed by researchers~\cite{madaan2023selfrefine,gou2024criticlargelanguagemodels,ozer2025marmultiagentreflexionimprovesreasoning,shinn2023reflexion}, rather than learning how humans naturally perform trial-and-error from data. Specifically, high-quality trial-and-error data should reflect two essential aspects: (1) multiple \emph{trials} at solving a task, and (2) evidence of learning from \emph{errors} across those trials. Existing datasets fail to capture these two dimensions simultaneously. For example, web agent benchmarks typically include only a single human trial per task, without recording repeated trials or feedback from errors~\cite{deng2024mind2web,zhou2024webarena,mialon2023gaia,wei2025browsecomp,he2024webvoyager,wu2025webwalker}. Other datasets, such as session search logs, may contain multiple trials, but they do not document how humans interpret errors and adjust their strategies accordingly~\cite{carterette2016trec, yandex2014,10.1145/3357384.3358158,10.1145/2911451.2914675,10.1145/3404835.3463242}.

To bridge this gap, we build an open-source platform \ours{} (\textbf{T}rial-and-\textbf{E}rror \textbf{C}ollection) for recording and annotating human trial-and-error behavior in web-based problem solving. The platform provides a Chrome extension that captures human behavioral trajectories on \emph{any} website non-intrusively, a Django system managing the full study lifecycle, and a replay-based annotation workflow that collects structured error diagnoses and corrective plans after each failed trial. Together, these components support both dimensions:  iterative \emph{trial} capture through cross-domain recording, and structured \emph{error} feedback through replay-based reflection.

Using this platform, we recruited 46 participants to collect the \ours{} dataset across 58 open-domain question-answering tasks. Participants tried to answer each question iteratively and produced 5,370 trials covering 41,229 webpages. Each trial record includes per-trial correctness labels, evidence markers, and full behavioral traces, satisfying the \emph{trial} dimension. Crucially, every failed trial is linked to a structured reflection consisting of a prioritized error diagnosis and corrective plan, capturing the \emph{error} dimension.

These data demonstrate that humans are highly efficient at trial-and-error. In contrast, current LLMs are substantially weaker: the best model matches human first-trial accuracy on \ours{}, but humans show significant accuracy gains across subsequent trials, whereas LLMs do not. Humans progressively shift their search strategies after error, while LLMs remain anchored to surface-level rephrasing. These results imply that \ours{} can serve as a valuable resource for developing LLM trial-and-error capabilities.

In summary, our contributions are: (1) the \ours{} platform, a system that records behavioral trajectories across iterative trials and collects error reflection through replay-based annotation; (2) the \ours{} dataset, the first collection built with this platform, comprising 5,370 trials with per-trial labels, behavioral traces, and error reflections across 58 information-seeking tasks; and (3) an illustrative analysis showing that the \ours{} dataset reveals concrete gaps between human and LLM trial-and-error capabilities on these tasks.

\section{Related Work}

Our work builds on two lines of prior research: web-based task resources and web interaction data collection platforms.

\subsection{Web-Based Task Resources}

\begin{table*}[htbp]
\caption{Comparison with existing resources along the dimensions of \emph{trial} and \emph{error}. The \emph{trial} dimension captures whether multiple trials are recorded with behavioral trajectories; the \emph{error} dimension captures whether per-trial correctness and error reflection are provided. \ours{} is the first resource satisfying both dimensions.}
\label{tab:comparison}
\small
\begin{tabular}{lccccc}
\toprule
& & \multicolumn{2}{c}{\textbf{\emph{Trial} Dimension}} & \multicolumn{2}{c}{\textbf{\emph{Error} Dimension}} \\
\cmidrule(lr){3-4} \cmidrule(lr){5-6}
\textbf{Resource} & \textbf{Task Type} & \textbf{Trials} & \textbf{Behavioral Traces} & \textbf{Explicit Feedback} & \textbf{Error Reflection}\\
\midrule
\multicolumn{6}{l}{\textit{Web agent benchmarks}} \\
Mind2Web & Web interaction & Single & \ding{51} (DOM actions) & \ding{51} (task success) & \ding{55} \\
WebArena & Web interaction & Single & \ding{55} & \ding{51} (task success) & \ding{55} \\
WebVoyager & Web interaction & Single & \ding{51} (screenshots) & \ding{51} (task success) & \ding{55} \\
\midrule
\multicolumn{6}{l}{\textit{Information seeking benchmarks}} \\
GAIA & Information seeking & Single & \ding{55} & \ding{51} (answer match) & \ding{55} \\
WebWalkerQA & Information seeking & Single & \ding{55} & \ding{51} (answer match) & \ding{55} \\
BrowseComp & Information seeking & Single & \ding{55} & \ding{51} (answer match) & \ding{55} \\
\midrule
\multicolumn{6}{l}{\textit{Session search logs}} \\
TREC Session Track & Information seeking & Multi-query & \ding{55} & \ding{55} (relevance, post-hoc) & \ding{55} \\
Yandex Personalized & Information seeking & Multi-query & \ding{55} & \ding{55} (clicks only) & \ding{55} \\
TianGong-ST & Information seeking & Multi-query & \ding{55} & \ding{55} (relevance, post-hoc) & \ding{55} \\
\midrule
\textbf{\ours{}} & \textbf{Information seeking} & \textbf{Multi-trial} & \textbf{\ding{51}} & \textbf{\ding{51} (per-trial labels)} & \textbf{\ding{51}} \\
\bottomrule
\end{tabular}
\end{table*}

Existing resources for web-based tasks fall into two broad families, as compared in Table~\ref{tab:comparison}. Web agent benchmarks provide interactive environments or evaluate answer correctness, but support only single trials without multi-trial behavioral traces, leaving the \emph{trial} dimension unaddressed. Web interaction benchmarks such as WebArena~\cite{zhou2024webarena} evaluate task success, and information seeking benchmarks such as BrowseComp~\cite{wei2025browsecomp} evaluate answer correctness, yet none supports iterative trials. Session search logs like TianGong-ST~\cite{10.1145/3357384.3358158} capture multi-query sequences with implicit signals (clicks, dwell times), but sessions are not segmented into discrete trials with per-trial labels, and none provides annotated reflections, leaving the \emph{error} dimension unaddressed. \ours{} is the first to satisfy both dimensions: multi-trial trajectories with behavioral traces, per-trial correctness feedback, and error reflection annotations.

\subsection{Web Interaction Data Collection Platforms}

Research logging frameworks~\cite{logui2021,10.1145/2910896.2925447,bhattacharya2021yasbil} capture interaction events but either require per-site configuration or lack replay and structured annotation. Commercial replay tools (e.g., OpenReplay) provide DOM replay but target product analytics, with no task management or reflection workflows. In contrast, \ours{}'s Chrome extension records trajectories on any website without per-site configuration, and the backend integrates replay with multi-trial task management (\emph{trial} dimension) and error reflections (\emph{error} dimension).

\section{\ours{} Platform}
\label{sec:platform}

\begin{figure}[htbp]
    \centering
    \includegraphics[width=\columnwidth]{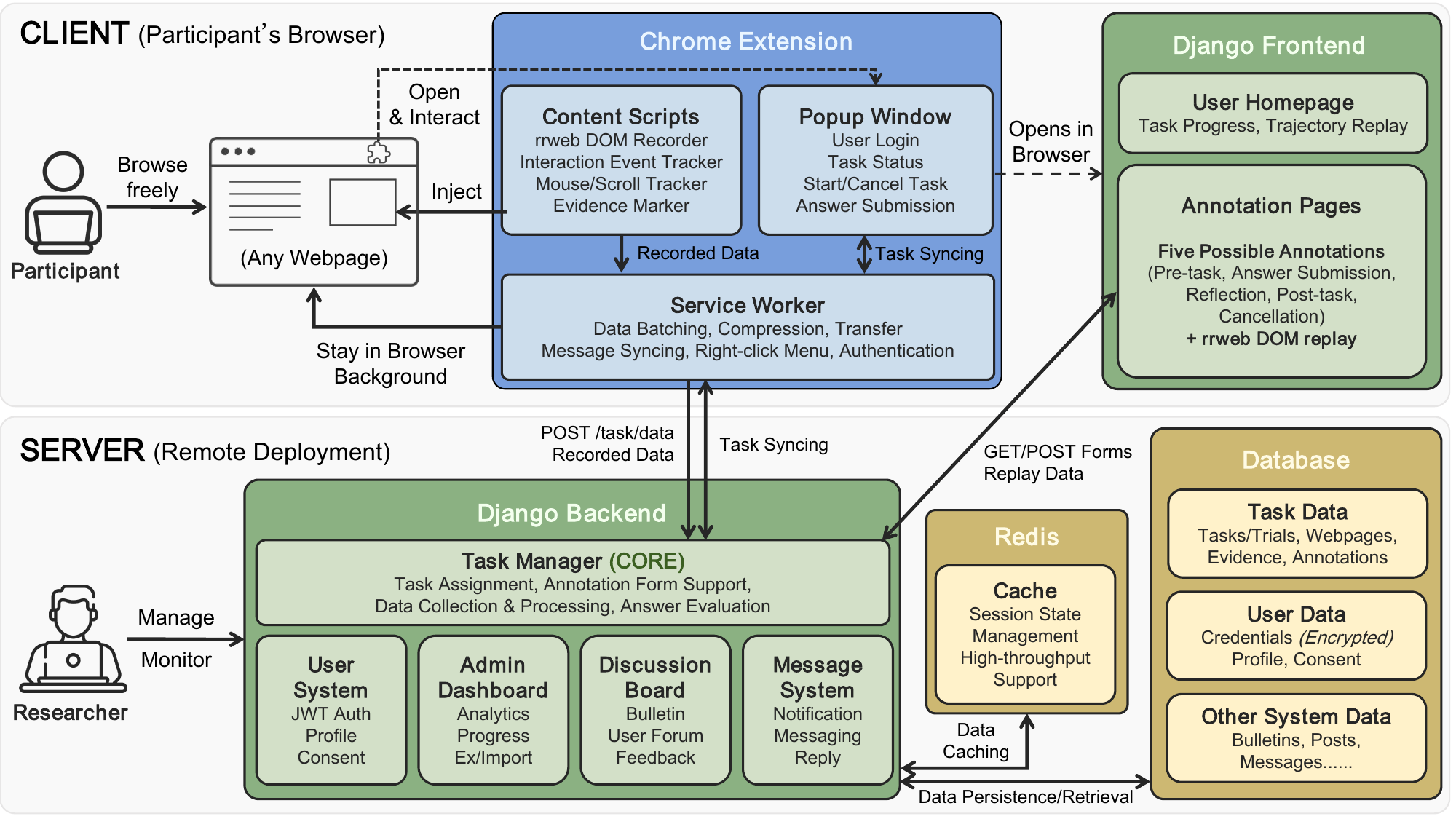}
    \caption{Platform architecture. The Chrome extension captures browsing trajectories from any webpage; the Django frontend supports annotation with replay; and the backend manages task assignment, data ingestion, and evaluation.}
    \label{fig:architecture}
\end{figure}

Guided by the two requirements identified above, the \ours{} platform (Figure~\ref{fig:architecture}) provides end-to-end infrastructure for web interaction studies in a single deployable system. The system architecture supports the \emph{trial} dimension by recording human behavioral traces and managing iterative task assignment with per-trial correctness feedback. The multi-stage annotation workflow built on top of it supports the \emph{error} dimension by eliciting structured error diagnoses and corrective plans through replay-based reflection.

\begin{figure}[htbp]
    \centering
    \includegraphics[width=0.8\columnwidth]{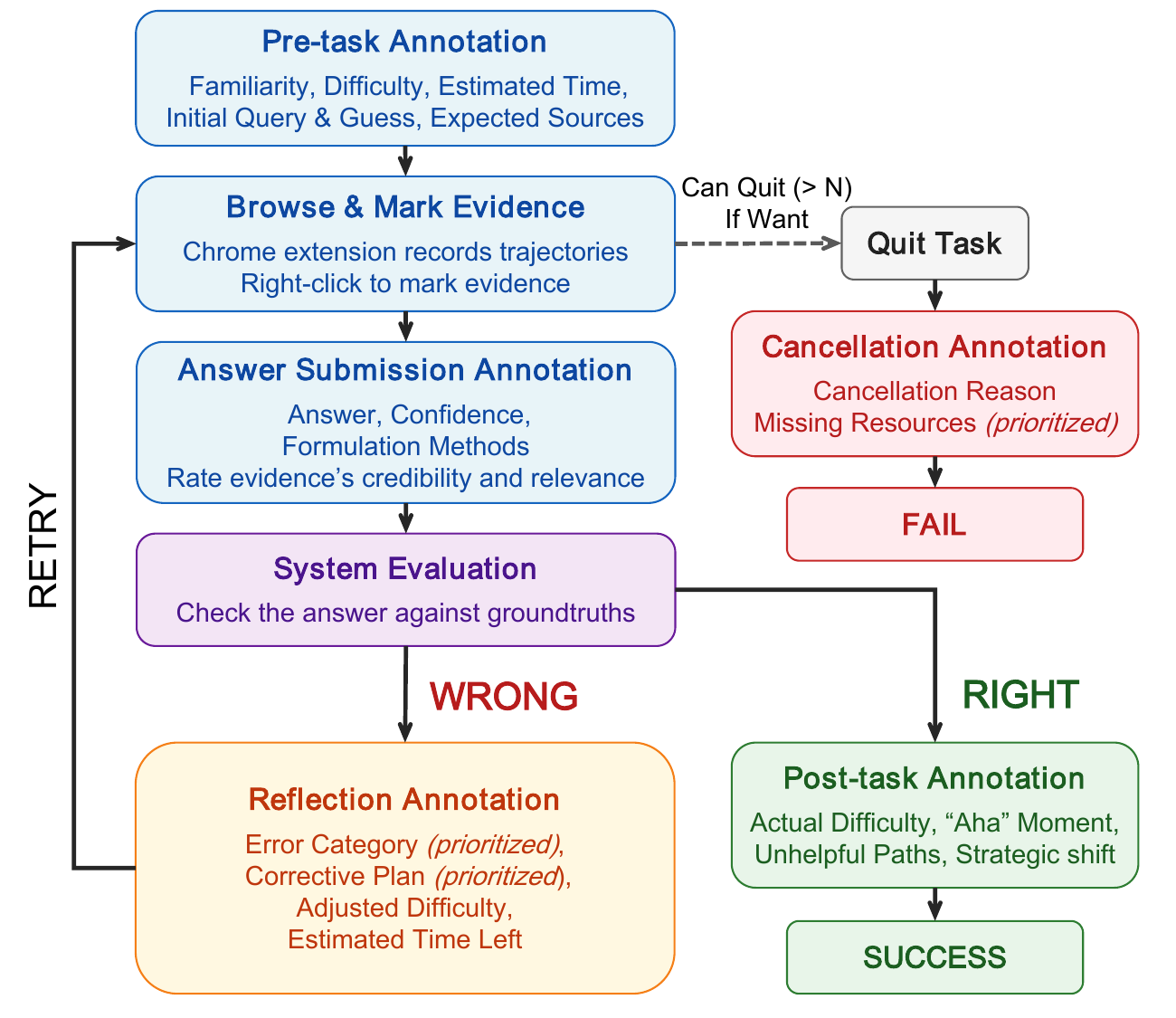}
    \caption{Multi-stage replay-based annotation workflow. On error, a reflection precedes the next trial; on success or cancellation, a corresponding assessment captures the outcome.}
    \label{fig:workflow}
\end{figure}

\subsection{System Architecture}
\label{sec:architecture}

\subsubsection{Chrome Extension}
Unlike platforms that instrument specific websites~\cite{deng2024mind2web, zhou2024webarena}, our extension works on \emph{any} webpage without per-site configuration. It is built on Chrome's current extension standard\footnote{\url{https://developer.chrome.com/docs/extensions/develop/migrate/what-is-mv3}} for long-term compatibility, whereas many existing tools still rely on the deprecated standard~\cite{10.1145/3442381.3450127,bhattacharya2021yasbil,10.1145/3411764.3445618}. For each page a user visits during a trial, the extension records three synchronized data streams: (1)~a replayable copy of the page via rrweb\footnote{\url{https://github.com/rrweb-io/rrweb}}, which captures the full page layout and all subsequent visual changes, enabling faithful replay without pixel-level screenshots at significantly lower storage cost; (2)~interaction events (clicks, hovers, key-presses, etc.) with timestamps and element identifiers; and (3)~continuous mouse position and scroll offset for reading pattern analysis. Page-level metadata (title, dwell time, referrer URL) is recorded automatically. Users can also mark evidence that supports their answer via a right-click menu, capturing the selected text, its position on the page, and the source URL. Recording is limited to designated task sessions for privacy concerns, and password fields are excluded by default.

\subsubsection{Backend System}

The backend manages the full study lifecycle: a core \textbf{task manager} organizes studies hierarchically (question $\rightarrow$ per-user task $\rightarrow$ trials) and handles task assignment, data collection, and answer evaluation with per-trial error feedback; a \textbf{user system} handles authentication, profiling, and versioned informed consent; and an \textbf{admin dashboard} provides real-time analytics, progress monitoring, and data export/import.

\subsection{Multi-Stage Replay-Based Annotation Workflow}
\label{sec:annotation}

\begin{figure*}[htbp]
    \centering
    \begin{subfigure}[b]{0.44\textwidth}
        \centering
        \includegraphics[width=\textwidth]{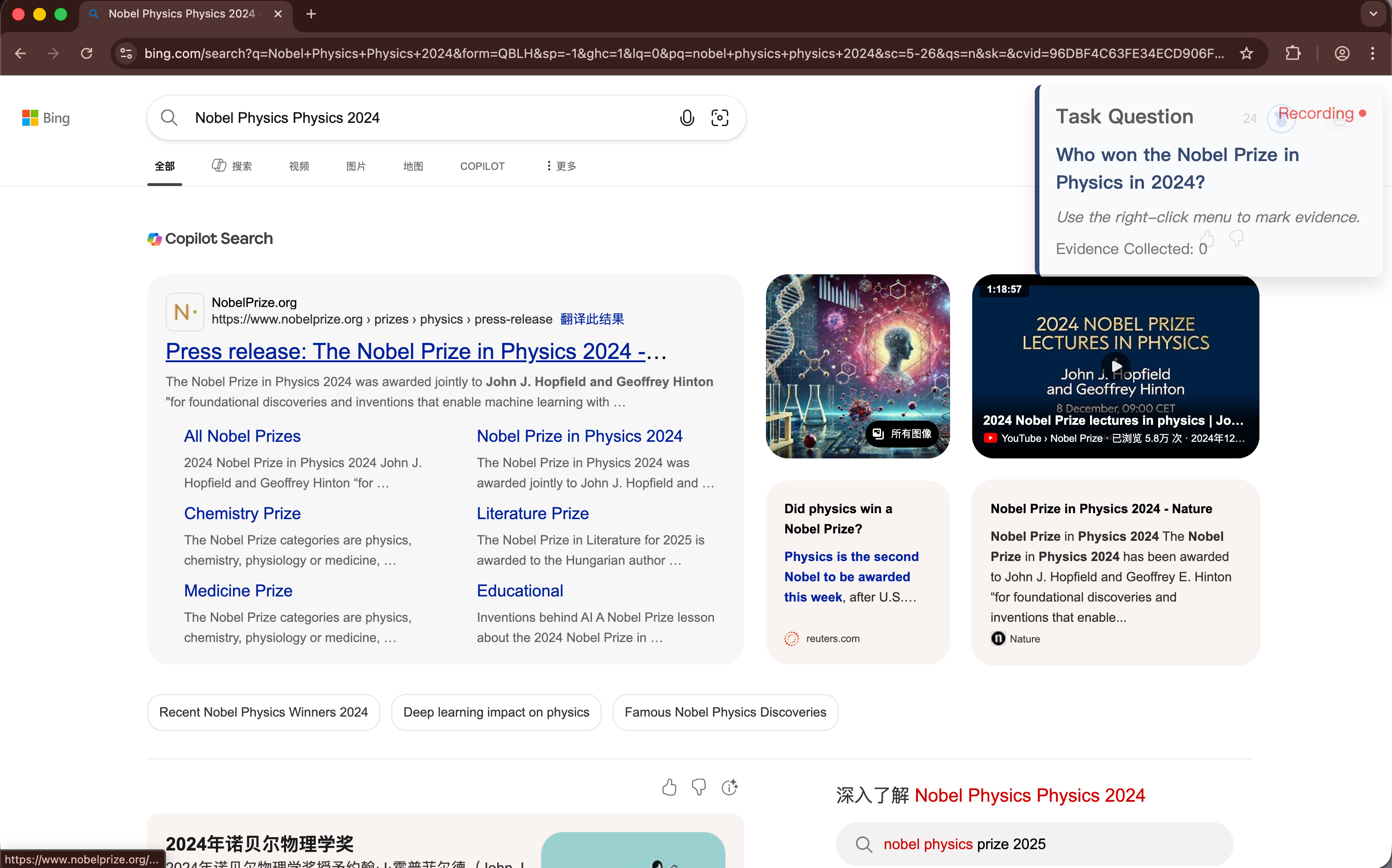}
        \caption{Browsing with extension}
        \label{fig:ss_browse}
    \end{subfigure}
    \quad
    \begin{subfigure}[b]{0.44\textwidth}
        \centering
        \includegraphics[width=\textwidth]{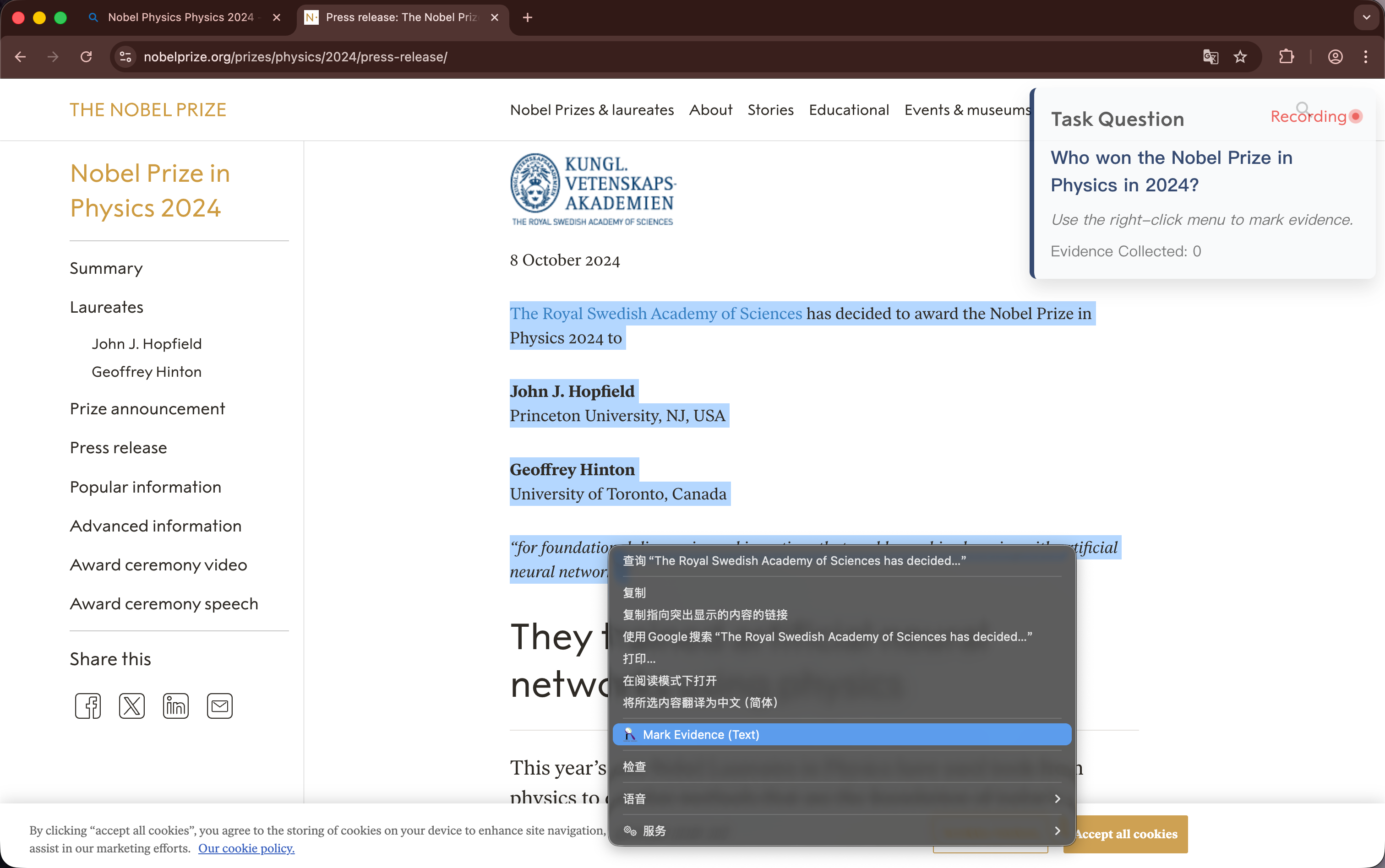}
        \caption{Evidence marking}
        \label{fig:ss_evidence}
    \end{subfigure}
    \vspace{0.5em}
    \begin{subfigure}[b]{0.44\textwidth}
        \centering
        \includegraphics[width=\textwidth]{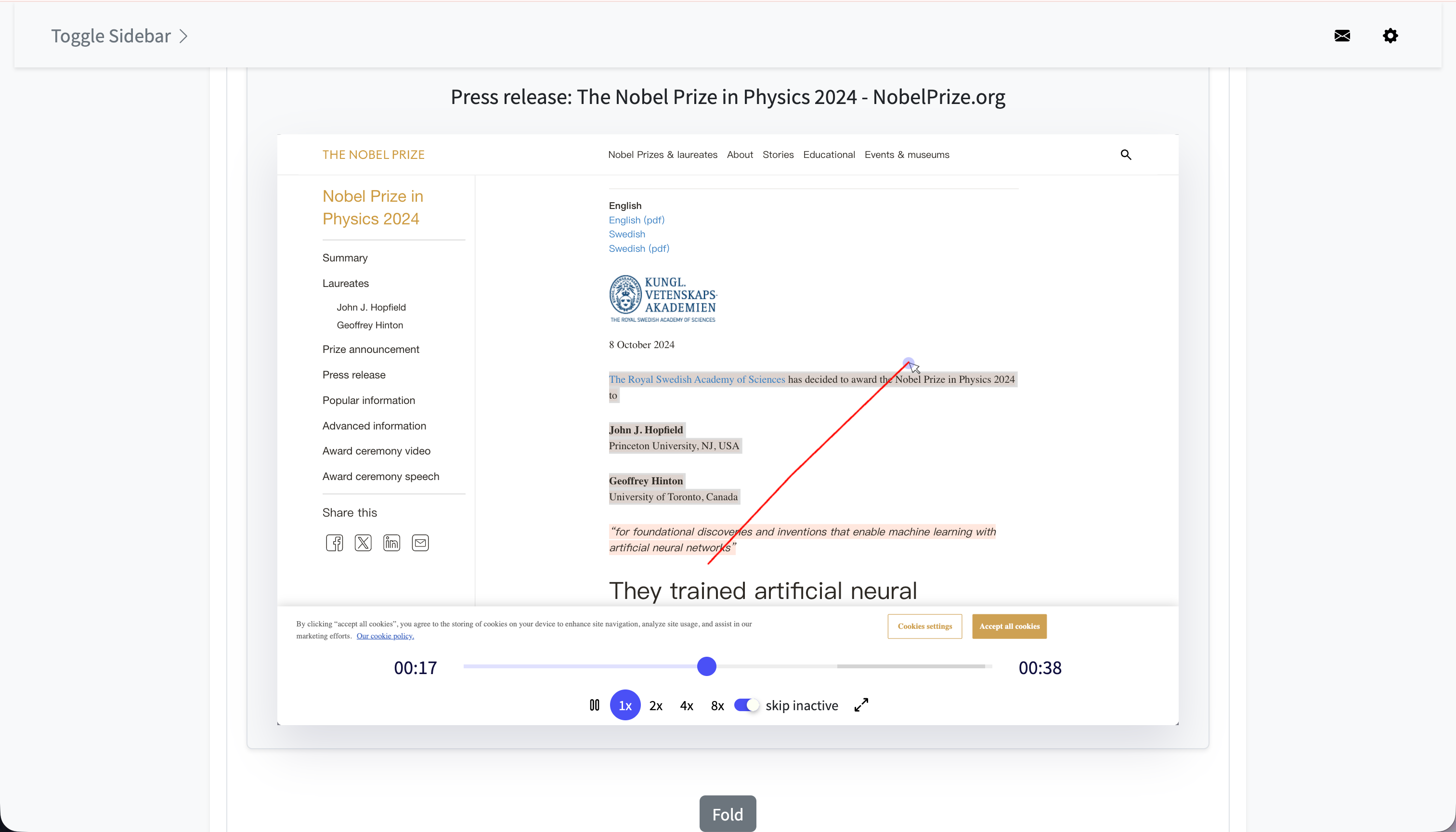}
        \caption{Replay viewer}
        \label{fig:ss_replay}
    \end{subfigure}
    \quad
    \begin{subfigure}[b]{0.44\textwidth}
        \centering
        \includegraphics[width=\textwidth]{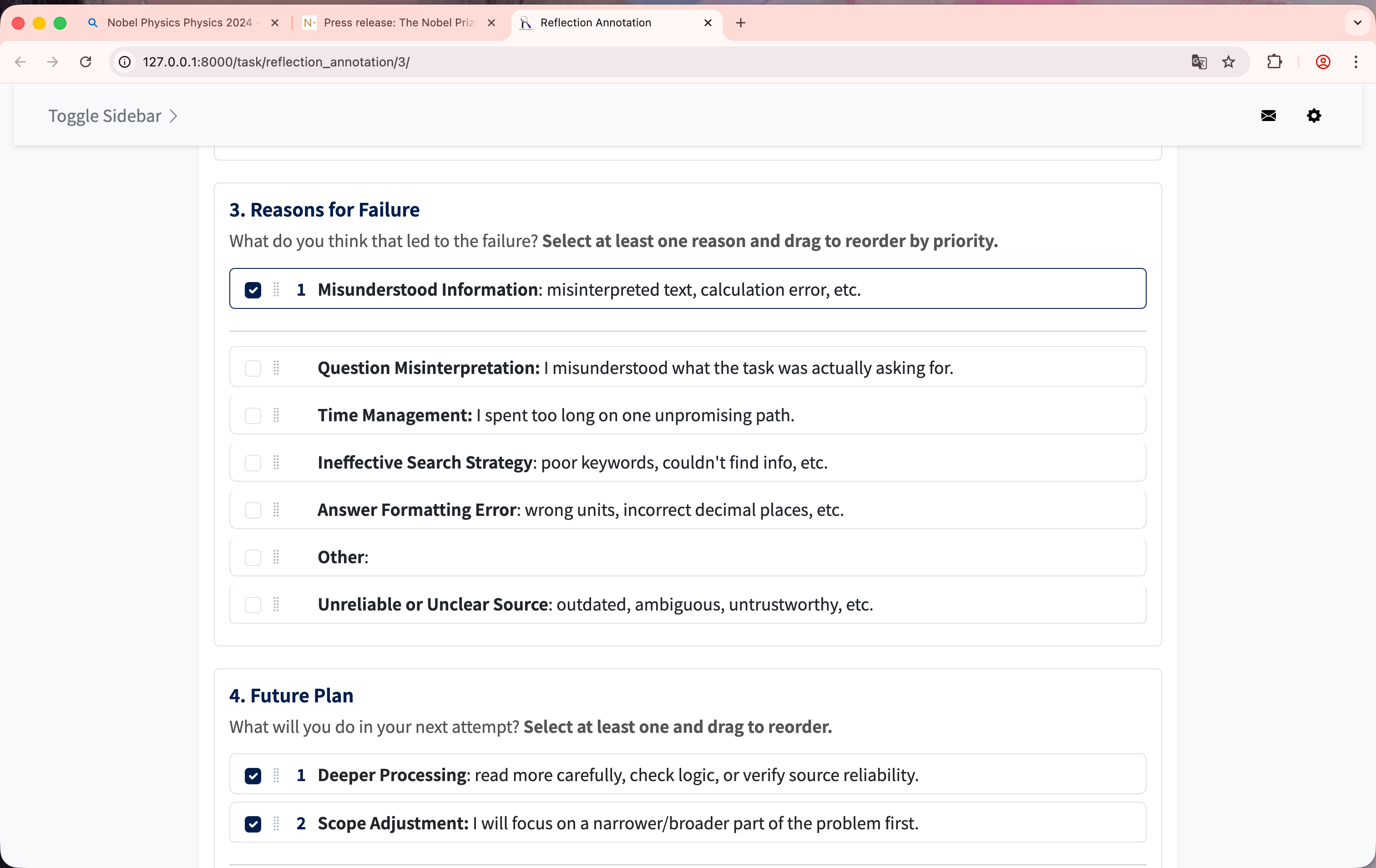}
        \caption{Reflection form}
        \label{fig:ss_reflection}
    \end{subfigure}
    \caption{Platform interfaces for multi-stage replay-based annotation workflow. (a) Participants browse freely while the extension records in the background; (b) evidence marking via right-click menu; (c) DOM-based replay of human trajectories and page changes; (d) reflection form collecting prioritized error diagnosis and corrective plan.}
    \label{fig:screenshots}
\end{figure*}

To capture error feedback without disrupting natural problem-solving behavior, users review a faithful replay of their own browsing session upon error and provide error diagnoses and corrective plans before retrying. This workflow (Figure~\ref{fig:workflow}) proceeds through five stages: (1) Pre-task assessment of familiarity, difficulty, and initial strategy. (2) Browse and collect evidence with extension recording; evidence marked via right-click menu. After N failed trials, participants can quit with a cancellation annotation. (3) Submit answer with confidence rating and evidence assessments. (4) Correctness evaluation against ground truth, routing to post-task assessment on success or reflection on error. (5) Reflection (on error) with prioritized diagnosis and corrective plan, then retry from stage (2). Figure~\ref{fig:screenshots} shows the platform interface. This replay-based annotation workflow can generalize to any web interaction task.

\section{\ours{} Dataset}
\label{sec:dataset}

Using the platform described in section \ref{sec:platform}, we collected a dataset of 5,370 trials across 58 questions. This section describes the data construction process, schema, and key characteristics.

\subsection{Data Construction}
\label{sec:protocol}

\textbf{Questions.} We sourced 58 open-domain factoid questions from Kamalloo et al.\cite{kamalloo2023evaluatingopendomainquestionanswering} where WebGPT~\cite{nakano2021webgpt} originally failed, as questions that LLMs already answer correctly offer limited insight into investigating their trial-and-error behavior. We removed ambiguous or disputed items, added temporal anchors for time-sensitive questions, and expanded answer variants for robust matching. 

\begin{figure}[htbp]
    \centering
    \begin{subfigure}[b]{0.46\columnwidth}
    \centering
    \includegraphics[width=\textwidth]{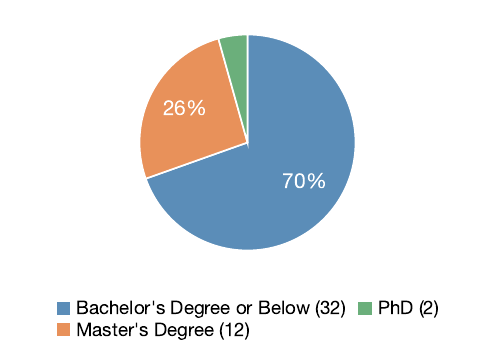}
    \caption{Educational Background}
    \label{fig:demo_education}
    \end{subfigure}
    \quad
    \begin{subfigure}[b]{0.46\columnwidth}
        \centering
        \includegraphics[width=\textwidth]{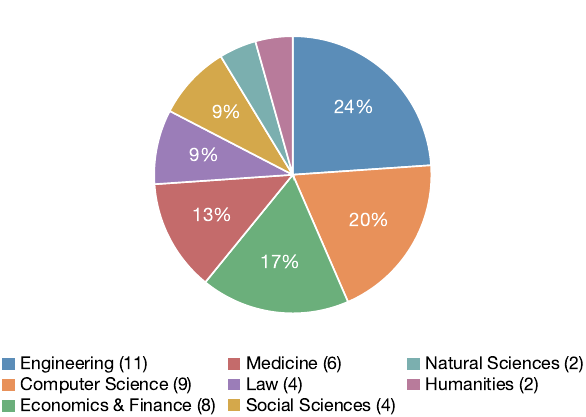}
        \caption{Academic Background}
        \label{fig:demo_field}
    \end{subfigure}
    \vspace{0.4em}
    \label{fig:distributions}
    \begin{subfigure}[b]{0.46\columnwidth}
        \centering
        \includegraphics[width=\textwidth]{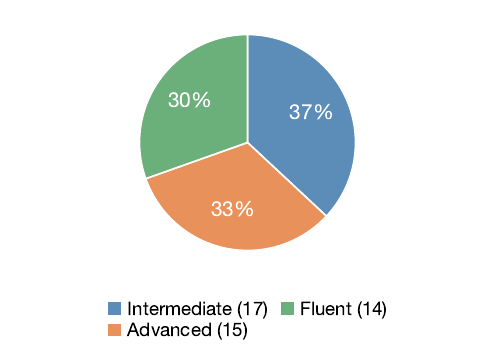}
        \caption{English Proficiency}
        \label{fig:demo_english}
    \end{subfigure}
    \quad
    \begin{subfigure}[b]{0.46\columnwidth}
        \centering
        \includegraphics[width=\textwidth]{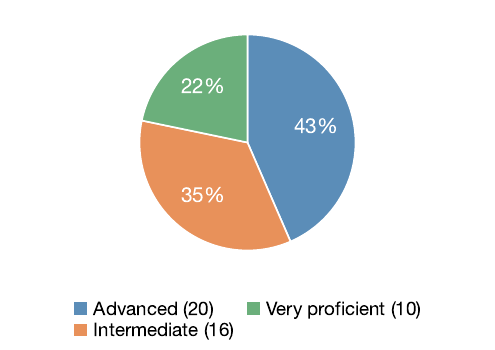}
        \caption{Web Search Proficiency}
        \label{fig:demo_search_prof}
    \end{subfigure}
    \caption{Demographic profile of 46 participants.}
    \label{fig:demo}
\end{figure}

\textbf{Participants and Protocol.} We recruited 46 participants from diverse fields via an online survey with English proficiency requirements and compensation at \$10 USD/hour. As shown in Figure~\ref{fig:demo}, participants were predominantly university students from diverse fields, with most reporting good english and web search proficiency. Each participant completed 4 tutorial questions before the formal study, and used an isolated browser profile without prior personal data for privacy. They tried iteratively and could give up after 5 unsuccessful trials. At least one evidence marker per submission was required. Questions appeared in randomized order. Each question received annotations from 42 participants on average. 

\textbf{Answer evaluation.}
\label{sec:evaluation}
Correctness is judged by GPT-4o based on the question, ground truths, and the participant's response. This LLM judge agrees with exact match in 94.8\% of trials (Cohen's $\kappa$=0.892) with disagreements primarily limited to paraphrasing.

\textbf{Data Postprocessing.}
We applied anomaly detection at the task, interaction, and user levels. Flagged trajectories and relevant annotations were manually reviewed and then removed. 

\textbf{Ethics and Licensing.} This study was approved by our Institutional Review Board. All participants signed informed consent. Browsing data was recorded only during designated task sessions, and the dataset is released with anonymized IDs and all private data removed.  All resources are released under the MIT license. Detailed documentation including data loading examples, file format specifications, and other tutorials are provided in the repository.

\subsection{Data Schema and Statistics}
\label{sec:schema}

\begin{table}[htbp]
\caption{Data schema of the \ours{} dataset.}
\label{tab:schema}
\small
\begin{tabular}{p{1.6cm}p{6.0cm}}
\toprule
\textbf{Record} & \textbf{Key Fields} \\
\midrule
\multicolumn{2}{l}{\textit{Per page}} \\
Webpage & URL, title, rrweb DOM recording, interaction events, mouse/scroll trajectory, dwell time, referrer \\
\midrule
\multicolumn{2}{l}{\textit{Per trial}} \\
Trial outcome & Answer, correctness, confidence, formulation method \\
Evidence & Selected text, DOM position, source URL, evidence type, relevance/credibility ratings \\
Reflection \newline {\scriptsize (on error)} & Error category, corrective plan (prioritized), adjusted difficulty, free-text notes \\
\midrule
\multicolumn{2}{l}{\textit{Per task}} \\
Pre-task & Familiarity, difficulty estimate, initial query plan, initial guess, expected sources \\
Post-task \newline {\scriptsize (on success)} & Actual difficulty, ``aha moment'' type, unhelpful paths, strategy shifts \\
Cancellation \newline {\scriptsize (on giving up)} & Cancellation reason, missing resources \\
\bottomrule
\end{tabular}
\end{table}

The dataset is organized around the trial-and-error loop, with its schema reflecting the two dimensions (Table~\ref{tab:schema}). Each task trajectory contains a pre-task assessment, one or more trials, and a terminal annotation (post-task on success, cancellation on giving up). For the \emph{trial} dimension, each trial includes page-level behavioral traces and an answer submission with evidence markers. For the \emph{error} dimension, per-trial correctness labels are provided, and every failed trial is linked to a reflection at trial $T$ that diagnoses why the trial failed and specifies a corrective plan for trial $T+1$.

\begin{table}[htbp]
\caption{Dataset statistics of \ours{}.}
\label{tab:statistics}
\small
\begin{tabular}{lr}
\toprule
\textbf{Metric} & \textbf{Value} \\
\midrule
\multicolumn{2}{l}{\textit{Scale}} \\
Questions & 58 \\
Tasks trajectories & 2,424 \\
Total trials & 5,370 \\
Avg trials per task & 2.2 \\
\midrule
\multicolumn{2}{l}{\textit{Browsing data}} \\
Webpages visited (with recording) &41,229 \\
Unique domains visited & 1,053 \\
Interaction & 1,516,981 \\
Search queries & 6,657 \\
\midrule
\multicolumn{2}{l}{\textit{Annotations}} \\
Pre-task assessments & 2,424 \\
Reflection annotations (error trial)& 2,946 \\
Post-task assessments (task succeeded)& 2,156 \\
Cancellation annotations (task failed)& 268 \\
Evidence markers & 7,208 \\
\bottomrule
\end{tabular}
\end{table}

Table~\ref{tab:statistics} summarizes the dataset. Of all task trajectories, 89\% end in success (with a post-task annotation) and 11\% in cancellation. Failed trials produce 2,946 reflections, each containing an error diagnosis and a corrective plan that bridges adjacent trials.

\begin{figure}[htbp]
    \centering
    \begin{subfigure}[b]{0.44\columnwidth}
        \centering
        \includegraphics[width=\textwidth]{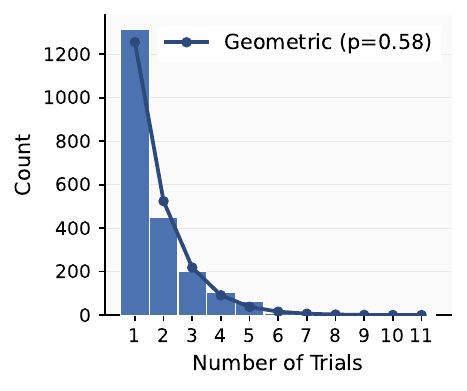}
        \caption{Trials to success}
        \label{fig:dist_trials}
    \end{subfigure}
    \quad
    \begin{subfigure}[b]{0.44\columnwidth}
        \centering
        \includegraphics[width=\textwidth]{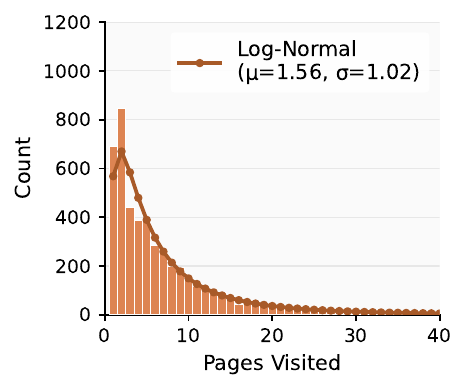}
        \caption{Pages per trial}
        \label{fig:dist_pages}
    \end{subfigure}
    \vspace{0.4em}
    \begin{subfigure}[b]{0.44\columnwidth}
        \centering
        \includegraphics[width=\textwidth]{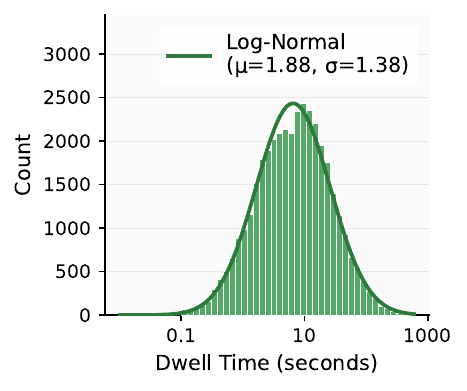}
        \caption{Dwell time per page}
        \label{fig:dist_dwell}
    \end{subfigure}
    \quad
    \begin{subfigure}[b]{0.44\columnwidth}
        \centering
        \includegraphics[width=\textwidth]{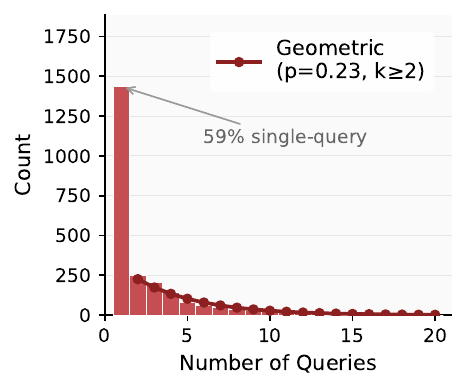}
        \caption{Queries per session}
        \label{fig:dist_queries}
    \end{subfigure}
    \caption{Distributions of four key behavioral dimensions.}
    \label{fig:distributions}
\end{figure}

Figure~\ref{fig:distributions} shows fitted distributions for four key behavioral dimensions. (a)~Trials to success follow a Geometric distribution, suggesting roughly constant per-trial success probability. (b--c)~Pages per trial and dwell time per page both follow heavy-tailed log-normal distributions, spanning a wide behavioral range. (d)~Queries per session exhibit a two-regime structure: 59\% of sessions resolve in a single query, while multi-query sessions follow a Geometric tail.

\subsection{Error Reflection Patterns}

\begin{figure}[htbp]
    \centering
    \includegraphics[width=0.67\columnwidth]{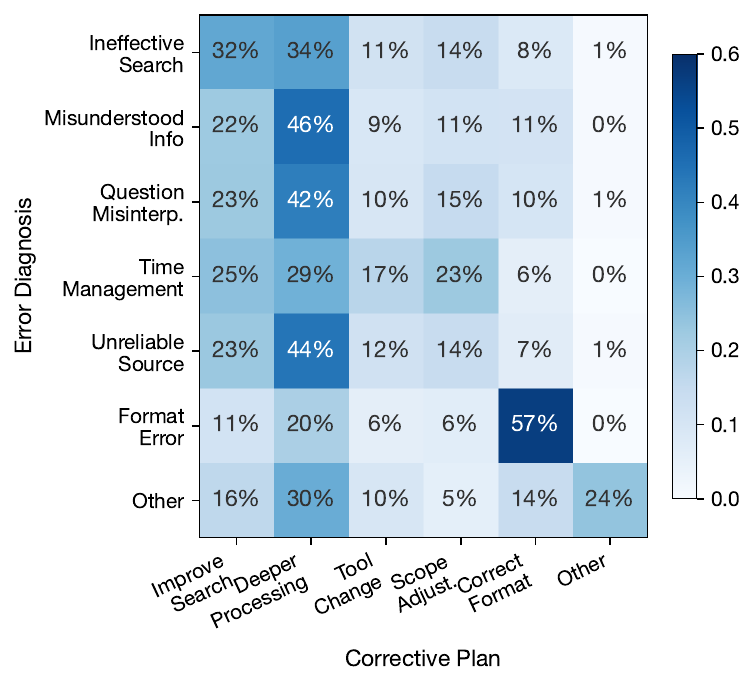}
    \caption{Conditional probability of corrective plan given error diagnosis, $P(\text{corrective plan} \mid \text{error diagnosis})$, over 2,946 reflections. Specific error triggers targeted strategies (e.g., ``Format Error'' $\rightarrow$ ``Correct Format'' at 57\%), showing that human reflection is diagnostic, not blind repetition.}
    \label{fig:reflection}
\end{figure}

Every failed trial includes a reflection consisting of an error diagnosis and a corrective plan. Figure~\ref{fig:reflection} shows the conditional distribution $P(\text{plan} \mid \text{diagnosis})$ over 2,946 reflections, revealing non-uniform mappings from diagnoses to plans. Some errors trigger focused strategies: ``Format Error'' leads to ``Correct Format'' in 57\% of cases, and ``Unreliable Source'' leads to ``Deeper Processing'' in 44\%. Others spread more broadly: ``Ineffective Search'' distributes across ``Improve Search'' (32\%) and ``Deeper Processing'' (34\%). This confirms that the \emph{error} feedback captured in \ours{} reflects genuine diagnostic reasoning rather than blind repetition. Do LLMs exhibit the same reflection behavior? We investigate this next.

\section{Illustrative Analysis: Human vs.\ LLM Comparison}
\label{sec:analysis}

To demonstrate what \ours{} uniquely enables, we compare human and LLM trial-and-error strategies on it. Because \ours{} captures both dimensions, we can assess not only whether LLMs match human accuracy on individual \emph{trials}, but also whether they recover effectively from \emph{errors} through reflection.

\subsection{Setup}

We compare four LLM baselines against humans:
(1)~\textbf{Vanilla LLM}: direct prompting without search.
(2)~\textbf{RAG}: query generation, search with full-text results~\cite{lewis2021retrievalaugmentedgenerationknowledgeintensivenlp},
and answer synthesis.
(3)~\textbf{Vanilla Agent}: a ReAct-style~\cite{yao2023reactsynergizingreasoningacting} agent with search and
page-visit tools that can selectively visit pages from search snippets.
(4)~\textbf{Browser Agent}: a ReAct-style agent with Chrome DevTools MCP\footnote{\url{https://github.com/ChromeDevTools/chrome-devtools-mcp}} that controls the browser as
humans do. Each baseline is run with GPT-4o-mini and Qwen3-8B across all 58 questions, with error reflection prompts between trials to match the human protocol. To ensure a fair comparison, all LLM baselines receive the full context of their prior trial before reflecting. In particular, Browser Agent observes the complete DOM tree and all page changes during browsing via Chrome DevTools MCP, providing retrospective context comparable to what human participants access through the replay interface. Vanilla Agent and RAG receive the full text of retrieved pages and their own action history.  All methods run for up to 5 trials; human trajectories are truncated at 5 for fair comparison.

\subsection{Performance Comparison}
\label{sec:reformulation}

\begin{figure*}[htbp]
    \centering
    \includegraphics[width=\textwidth]{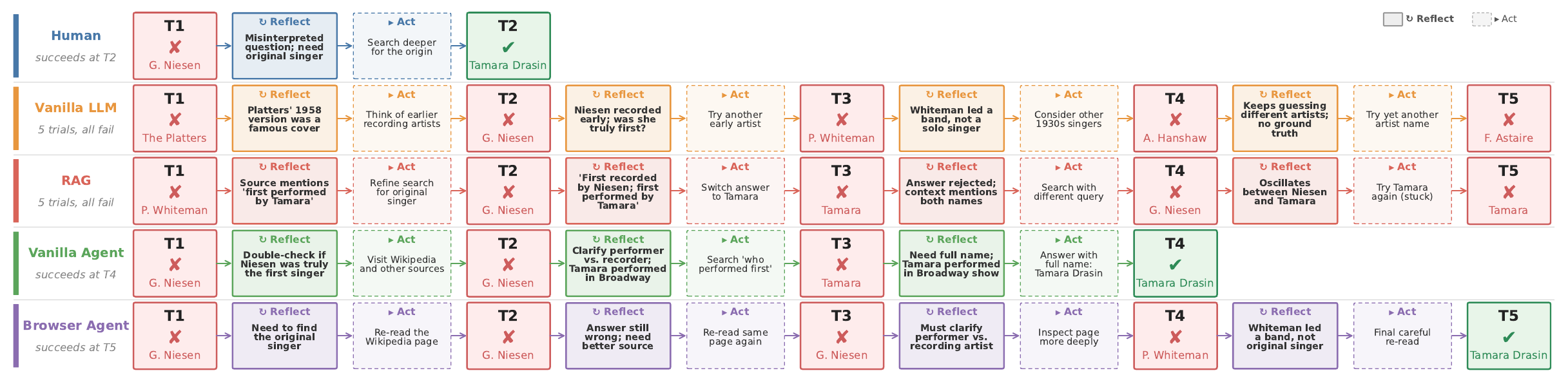}
    \caption{Case study: ``Who sang Smoke Gets in Your Eyes first?'' (answer:
Tamara Drasin). Each row shows one method's \emph{reflect-then-act} trial
sequence. The human diagnoses a misinterpretation and succeeds at T2;
Vanilla LLM and RAG both fail all 5 trials; Vanilla Agent succeeds by
T4 through iterative search; Browser Agent converges at T5 after
prolonged fixation.}
    \label{fig:casestudy}
\end{figure*}

\begin{table}[htbp]
\caption{Human vs.\ LLM performance on 58 questions (up to 5 trials). SR@$k$: fraction of tasks solved within $k$ trials. Recovery Rate: $P(\text{success} \leq 5 \mid \text{fail at T1})$, measuring the ability to succeed after an initial error. Avg~T: average number of trials used. The best LLM matches human first-trial accuracy but recovers from errors at a substantially lower rate.}
\label{tab:performance}
\footnotesize
\begin{tabular}{llrrrr}
\toprule
\textbf{Method} & \textbf{Model} & \textbf{SR@1} & \textbf{SR@5} & \textbf{Recovery} & \textbf{Avg T} \\
\midrule
\textbf{Human} & --- & \textbf{56.6} & \textbf{88.9} & \textbf{74.5} & \textbf{2.14} \\
\midrule
\multirow{2}{*}{Vanilla LLM} & GPT-4o-mini & 34.5 & 55.2 & 31.6 & 3.21 \\
 & Qwen3-8B & 21.4 & 32.1 & 13.6 & 3.89 \\
\multirow{2}{*}{RAG} & GPT-4o-mini & 50.0 & 72.4 & 44.8 & 2.38 \\
 & Qwen3-8B & 48.1 & 68.5 & 39.3 & 2.54 \\
\multirow{2}{*}{Vanilla Agent} & GPT-4o-mini & \textbf{58.6} & 79.3 & 50.0 & 2.17 \\
 & Qwen3-8B & 50.0 & 64.8 & 29.6 & 2.72 \\
\multirow{2}{*}{Browser Agent} & GPT-4o-mini & 36.8 & 59.6 & 37.1 & 2.63 \\
 & Qwen3-8B & 18.6 & 32.6 & 17.1 & 3.98 \\
\bottomrule
\end{tabular}
\end{table}

\begin{figure}[htbp]
    \centering
    \begin{subfigure}[b]{0.48\columnwidth}
        \centering
        \includegraphics[width=\textwidth]{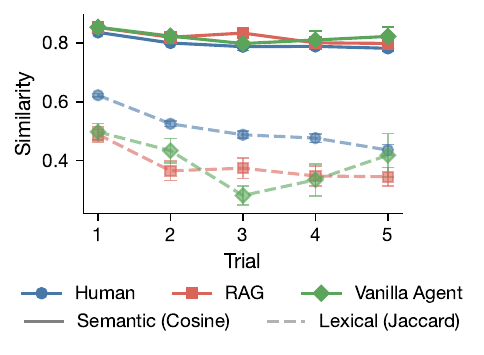}
        \caption{Similarity to question}
        \label{fig:reform_question}
    \end{subfigure}
    \hfill
    \begin{subfigure}[b]{0.48\columnwidth}
        \centering
        \includegraphics[width=\textwidth]{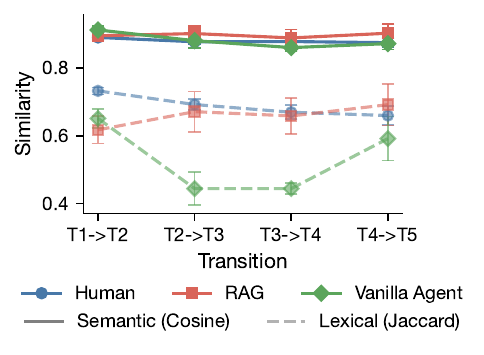}
        \caption{Inter-trial similarity}
        \label{fig:reform_prev}
    \end{subfigure}
    \caption{Query reformulation patterns (GPT-4o-mini). (a) Similarity between queries and original question; (b) pairwise similarity between queries in adjacent trials. We exclude Browser Agent and Vanilla LLM here as they submit few or no search queries. Humans progressively diverge in semantic space, while LLMs show only surface-level lexical changes.}
    \label{fig:reformulation}
\end{figure}



Table~\ref{tab:performance} summarizes the experimental results. On the \emph{trial} dimension, the best LLM (Vanilla Agent, GPT-4o-mini) achieves comparable first-trial accuracy (SR@1). However, on the \emph{error} dimension, the recovery rate gap is stark: 50.0\% for the best LLM vs.\ 74.5\% for humans, indicating that LLMs struggle to learn from errors. Notably, Browser Agent underperforms all other baselines despite having the richest tool set, as it over-relies on parametric knowledge to navigate URLs directly rather than querying search engines. Furthermore, the cross-trial data in \ours{} reveals what drives this gap in the \emph{error} dimension. Since each trial may contain multiple search queries, we measure inter-trial query similarity as the mean pairwise similarity between all queries in adjacent trials, using Qwen3-Embedding-0.6B~\cite{zhang2025qwen3embeddingadvancingtext} for semantic similarity and Jaccard overlap for lexical similarity. Figure~\ref{fig:reformulation} shows that human queries progressively diverge from the original question in semantic space, indicating genuine strategy shifts after error. LLM queries, by contrast, remain anchored with only surface-level lexical changes, suggesting that their reflections fail to produce substantive reformulations.

\subsection{Case Study}
\label{sec:casestudy}

Figure~\ref{fig:casestudy} traces all five methods on ``Who sang Smoke Gets in Your Eyes first?'' (answer: Tamara Drasin), where the difficulty lies in distinguishing the original 1933 Broadway performer from other recording artists such as Gertrude Niesen. The human answers ``G.~Niesen'' at T1, reflects diagnostically (\emph{question misinterpretation} $\rightarrow$ \emph{search deeper}), and succeeds at T2, exemplifying effective use of the \emph{error} signal. Both Vanilla LLM and RAG fail all 5 trials: Vanilla LLM cycles through artist names without external grounding, while RAG retrieves context mentioning ``Tamara'' yet oscillates between Niesen and Tamara and keeps missing the surname ``Drasin.'' This near-miss phenomenon shows that retrieval can bring a model close yet fail without precise reflection to complete the last step. Vanilla Agent succeeds by T4 through iterative web search, and Browser Agent fixates on Niesen for three trials before converging at T5.

\section{Conclusion and Future Work}

We present \ours{}, an open-source platform and dataset for studying human trial-and-error in problem solving with web search. The platform is designed around the two dimensions of \emph{trial} and \emph{error}. It supports iterative trial capture through behavioral trajectory recording and structured error feedback through a replay-based annotation workflow. The resulting dataset of 5,370 trials links multi-trial trajectories with per-trial labels, diagnostic reflections, and behavioral traces. Our comparison of four LLM baselines against humans shows that the best model matches human first-trial accuracy, but humans show significant accuracy gains across subsequent trials, whereas LLMs do not. Humans progressively shift their search queries after error, while LLMs remain anchored to surface-level rephrasing. These results imply that \ours{} can offer valuable data for developing LLM trial-and-error capabilities.

For future work, we plan to pursue two directions. First, we will use the \ours{} platform to collect more data, extending to decision-making and open-ended exploration tasks to broaden both the data types and scale. Second, we will leverage the collected data to improve LLMs, including training agents with annotated human reflection feedback and building realistic multi-trial user simulators.

\begin{acks}
    This work is supported by the Research Project of Quancheng Laboratory, China (Grant No. QCL20250105) and Key R\&D Program of Shandong Province (SYS202201)
\end{acks}

\bibliographystyle{ACM-Reference-Format}
\bibliography{sample-base}.

@inproceedings{shinn2023reflexion,
author = {Shinn, Noah and Cassano, Federico and Gopinath, Ashwin and Narasimhan, Karthik and Yao, Shunyu},
title = {Reflexion: language agents with verbal reinforcement learning},
year = {2023},
publisher = {Curran Associates Inc.},
address = {Red Hook, NY, USA},
booktitle = {Proceedings of the 37th International Conference on Neural Information Processing Systems},
articleno = {377},
numpages = {19},
location = {New Orleans, LA, USA},
series = {NIPS '23}
}

@inproceedings{madaan2023selfrefine,
author = {Madaan, Aman and Tandon, Niket and Gupta, Prakhar and Hallinan, Skyler and Gao, Luyu and Wiegreffe, Sarah and Alon, Uri and Dziri, Nouha and Prabhumoye, Shrimai and Yang, Yiming and Gupta, Shashank and Majumder, Bodhisattwa Prasad and Hermann, Katherine and Welleck, Sean and Yazdanbakhsh, Amir and Clark, Peter},
title = {SELF-REFINE: iterative refinement with self-feedback},
year = {2023},
publisher = {Curran Associates Inc.},
address = {Red Hook, NY, USA},
booktitle = {Proceedings of the 37th International Conference on Neural Information Processing Systems},
articleno = {2019},
numpages = {61},
location = {New Orleans, LA, USA},
series = {NIPS '23}
}

@inproceedings{carterette2016trec,
  author    = {Ben Carterette and
               Evangelos Kanoulas and
               Mark M. Hall and
               Paul D. Clough},
  editor    = {Ellen M. Voorhees and
               Angela Ellis},
  title     = {Overview of the {TREC} 2014 Session Track},
  booktitle = {Proceedings of The Twenty-Third Text REtrieval Conference, {TREC}
               2014, Gaithersburg, Maryland, USA, November 19-21, 2014},
  series    = {{NIST} Special Publication},
  volume    = {500-308},
  publisher = {National Institute of Standards and Technology {(NIST)}},
  year      = {2014},
  url       = {http://trec.nist.gov/pubs/trec23/papers/overview-session.pdf},
  bibsource = {dblp computer science bibliography, https://dblp.org}
}

@inproceedings{10.1145/2911451.2914675,
author = {Carterette, Ben and Clough, Paul and Hall, Mark and Kanoulas, Evangelos and Sanderson, Mark},
title = {Evaluating Retrieval over Sessions: The TREC Session Track 2011-2014},
year = {2016},
isbn = {9781450340694},
publisher = {Association for Computing Machinery},
address = {New York, NY, USA},
url = {https://doi.org/10.1145/2911451.2914675},
doi = {10.1145/2911451.2914675},
booktitle = {Proceedings of the 39th International ACM SIGIR Conference on Research and Development in Information Retrieval},
pages = {685–688},
numpages = {4},
location = {Pisa, Italy},
series = {SIGIR '16}
}

@inproceedings{deng2024mind2web,
author = {Deng, Xiang and Gu, Yu and Zheng, Boyuan and Chen, Shijie and Stevens, Samuel and Wang, Boshi and Sun, Huan and Su, Yu},
title = {MIND2WEB: towards a generalist agent for the web},
year = {2023},
publisher = {Curran Associates Inc.},
address = {Red Hook, NY, USA},
booktitle = {Proceedings of the 37th International Conference on Neural Information Processing Systems},
articleno = {1220},
numpages = {24},
location = {New Orleans, LA, USA},
series = {NIPS '23}
}

@inproceedings{10.1145/3404835.3463242,
author = {Rekabsaz, Navid and Lesota, Oleg and Schedl, Markus and Brassey, Jon and Eickhoff, Carsten},
title = {TripClick: The Log Files of a Large Health Web Search Engine},
year = {2021},
isbn = {9781450380379},
publisher = {Association for Computing Machinery},
address = {New York, NY, USA},
url = {https://doi.org/10.1145/3404835.3463242},
doi = {10.1145/3404835.3463242},
booktitle = {Proceedings of the 44th International ACM SIGIR Conference on Research and Development in Information Retrieval},
pages = {2507–2513},
numpages = {7},
location = {Virtual Event, Canada},
series = {SIGIR '21}
}

@misc{zhou2024webarena,
      title={WebArena: A Realistic Web Environment for Building Autonomous Agents}, 
      author={Shuyan Zhou and Frank F. Xu and Hao Zhu and Xuhui Zhou and Robert Lo and Abishek Sridhar and Xianyi Cheng and Tianyue Ou and Yonatan Bisk and Daniel Fried and Uri Alon and Graham Neubig},
      year={2024},
      eprint={2307.13854},
      archivePrefix={arXiv},
      primaryClass={cs.AI},
      url={https://arxiv.org/abs/2307.13854}, 
}

@misc{wei2025browsecomp,
      title={BrowseComp: A Simple Yet Challenging Benchmark for Browsing Agents}, 
      author={Jason Wei and Zhiqing Sun and Spencer Papay and Scott McKinney and Jeffrey Han and Isa Fulford and Hyung Won Chung and Alex Tachard Passos and William Fedus and Amelia Glaese},
      year={2025},
      eprint={2504.12516},
      archivePrefix={arXiv},
      primaryClass={cs.CL},
      url={https://arxiv.org/abs/2504.12516}, 
}

@misc{mialon2023gaia,
      title={GAIA: a benchmark for General AI Assistants}, 
      author={Grégoire Mialon and Clémentine Fourrier and Craig Swift and Thomas Wolf and Yann LeCun and Thomas Scialom},
      year={2023},
      eprint={2311.12983},
      archivePrefix={arXiv},
      primaryClass={cs.CL},
      url={https://arxiv.org/abs/2311.12983}, 
}

@inproceedings{logui2021,
author = {Maxwell, David and Hauff, Claudia},
title = {LogUI: Contemporary Logging Infrastructure for Web-Based Experiments},
year = {2021},
isbn = {978-3-030-72239-5},
publisher = {Springer-Verlag},
address = {Berlin, Heidelberg},
url = {https://doi.org/10.1007/978-3-030-72240-1_59},
doi = {10.1007/978-3-030-72240-1_59},
booktitle = {Advances in  Information Retrieval: 43rd European Conference on IR Research, ECIR 2021, Virtual Event, March 28 – April 1, 2021, Proceedings, Part II},
pages = {525–530},
numpages = {6}
}

@misc{kamalloo2023evaluatingopendomainquestionanswering,
      title={Evaluating Open-Domain Question Answering in the Era of Large Language Models}, 
      author={Ehsan Kamalloo and Nouha Dziri and Charles L. A. Clarke and Davood Rafiei},
      year={2023},
      eprint={2305.06984},
      archivePrefix={arXiv},
      primaryClass={cs.CL},
      url={https://arxiv.org/abs/2305.06984}, 
}

@inproceedings{he2024webvoyager,
    title = "{W}eb{V}oyager: Building an End-to-End Web Agent with Large Multimodal Models",
    author = "He, Hongliang  and
      Yao, Wenlin  and
      Ma, Kaixin  and
      Yu, Wenhao  and
      Dai, Yong  and
      Zhang, Hongming  and
      Lan, Zhenzhong  and
      Yu, Dong",
    editor = "Ku, Lun-Wei  and
      Martins, Andre  and
      Srikumar, Vivek",
    booktitle = "Proceedings of the 62nd Annual Meeting of the Association for Computational Linguistics (Volume 1: Long Papers)",
    month = aug,
    year = "2024",
    address = "Bangkok, Thailand",
    publisher = "Association for Computational Linguistics",
    url = "https://aclanthology.org/2024.acl-long.371/",
    doi = "10.18653/v1/2024.acl-long.371",
    pages = "6864--6890"
}

@inproceedings{wu2025webwalker,
    title = "{W}eb{W}alker: Benchmarking {LLM}s in Web Traversal",
    author = "Wu, Jialong  and
      Yin, Wenbiao  and
      Jiang, Yong  and
      Wang, Zhenglin  and
      Xi, Zekun  and
      Fang, Runnan  and
      Zhang, Linhai  and
      He, Yulan  and
      Zhou, Deyu  and
      Xie, Pengjun  and
      Huang, Fei",
    editor = "Che, Wanxiang  and
      Nabende, Joyce  and
      Shutova, Ekaterina  and
      Pilehvar, Mohammad Taher",
    booktitle = "Proceedings of the 63rd Annual Meeting of the Association for Computational Linguistics (Volume 1: Long Papers)",
    month = jul,
    year = "2025",
    address = "Vienna, Austria",
    publisher = "Association for Computational Linguistics",
    url = "https://aclanthology.org/2025.acl-long.508/",
    doi = "10.18653/v1/2025.acl-long.508",
    pages = "10290--10305",
    ISBN = "979-8-89176-251-0"
}

@misc{nakano2021webgpt,
      title={WebGPT: Browser-assisted question-answering with human feedback}, 
      author={Reiichiro Nakano and Jacob Hilton and Suchir Balaji and Jeff Wu and Long Ouyang and Christina Kim and Christopher Hesse and Shantanu Jain and Vineet Kosaraju and William Saunders and Xu Jiang and Karl Cobbe and Tyna Eloundou and Gretchen Krueger and Kevin Button and Matthew Knight and Benjamin Chess and John Schulman},
      year={2022},
      eprint={2112.09332},
      archivePrefix={arXiv},
      primaryClass={cs.CL},
      url={https://arxiv.org/abs/2112.09332}, 
}

@article{Guo_2025,
   title={DeepSeek-R1 incentivizes reasoning in LLMs through reinforcement learning},
   volume={645},
   ISSN={1476-4687},
   url={http://dx.doi.org/10.1038/s41586-025-09422-z},
   DOI={10.1038/s41586-025-09422-z},
   number={8081},
   journal={Nature},
   publisher={Springer Science and Business Media LLC},
   author={Guo, Daya and Yang, Dejian and Zhang, Haowei and Song, Junxiao and Wang, Peiyi and Zhu, Qihao and Xu, Runxin and Zhang, Ruoyu and Ma, Shirong and Bi, Xiao and Zhang, Xiaokang and Yu, Xingkai and Wu, Yu and Wu, Z. F. and Gou, Zhibin and Shao, Zhihong and Li, Zhuoshu and Gao, Ziyi and Liu, Aixin and Xue, Bing and Wang, Bingxuan and Wu, Bochao and Feng, Bei and Lu, Chengda and Zhao, Chenggang and Deng, Chengqi and Ruan, Chong and Dai, Damai and Chen, Deli and Ji, Dongjie and Li, Erhang and Lin, Fangyun and Dai, Fucong and Luo, Fuli and Hao, Guangbo and Chen, Guanting and Li, Guowei and Zhang, H. and Xu, Hanwei and Ding, Honghui and Gao, Huazuo and Qu, Hui and Li, Hui and Guo, Jianzhong and Li, Jiashi and Chen, Jingchang and Yuan, Jingyang and Tu, Jinhao and Qiu, Junjie and Li, Junlong and Cai, J. L. and Ni, Jiaqi and Liang, Jian and Chen, Jin and Dong, Kai and Hu, Kai and You, Kaichao and Gao, Kaige and Guan, Kang and Huang, Kexin and Yu, Kuai and Wang, Lean and Zhang, Lecong and Zhao, Liang and Wang, Litong and Zhang, Liyue and Xu, Lei and Xia, Leyi and Zhang, Mingchuan and Zhang, Minghua and Tang, Minghui and Zhou, Mingxu and Li, Meng and Wang, Miaojun and Li, Mingming and Tian, Ning and Huang, Panpan and Zhang, Peng and Wang, Qiancheng and Chen, Qinyu and Du, Qiushi and Ge, Ruiqi and Zhang, Ruisong and Pan, Ruizhe and Wang, Runji and Chen, R. J. and Jin, R. L. and Chen, Ruyi and Lu, Shanghao and Zhou, Shangyan and Chen, Shanhuang and Ye, Shengfeng and Wang, Shiyu and Yu, Shuiping and Zhou, Shunfeng and Pan, Shuting and Li, S. S. and Zhou, Shuang and Wu, Shaoqing and Yun, Tao and Pei, Tian and Sun, Tianyu and Wang, T. and Zeng, Wangding and Liu, Wen and Liang, Wenfeng and Gao, Wenjun and Yu, Wenqin and Zhang, Wentao and Xiao, W. L. and An, Wei and Liu, Xiaodong and Wang, Xiaohan and Chen, Xiaokang and Nie, Xiaotao and Cheng, Xin and Liu, Xin and Xie, Xin and Liu, Xingchao and Yang, Xinyu and Li, Xinyuan and Su, Xuecheng and Lin, Xuheng and Li, X. Q. and Jin, Xiangyue and Shen, Xiaojin and Chen, Xiaosha and Sun, Xiaowen and Wang, Xiaoxiang and Song, Xinnan and Zhou, Xinyi and Wang, Xianzu and Shan, Xinxia and Li, Y. K. and Wang, Y. Q. and Wei, Y. X. and Zhang, Yang and Xu, Yanhong and Li, Yao and Zhao, Yao and Sun, Yaofeng and Wang, Yaohui and Yu, Yi and Zhang, Yichao and Shi, Yifan and Xiong, Yiliang and He, Ying and Piao, Yishi and Wang, Yisong and Tan, Yixuan and Ma, Yiyang and Liu, Yiyuan and Guo, Yongqiang and Ou, Yuan and Wang, Yuduan and Gong, Yue and Zou, Yuheng and He, Yujia and Xiong, Yunfan and Luo, Yuxiang and You, Yuxiang and Liu, Yuxuan and Zhou, Yuyang and Zhu, Y. X. and Huang, Yanping and Li, Yaohui and Zheng, Yi and Zhu, Yuchen and Ma, Yunxian and Tang, Ying and Zha, Yukun and Yan, Yuting and Ren, Z. Z. and Ren, Zehui and Sha, Zhangli and Fu, Zhe and Xu, Zhean and Xie, Zhenda and Zhang, Zhengyan and Hao, Zhewen and Ma, Zhicheng and Yan, Zhigang and Wu, Zhiyu and Gu, Zihui and Zhu, Zijia and Liu, Zijun and Li, Zilin and Xie, Ziwei and Song, Ziyang and Pan, Zizheng and Huang, Zhen and Xu, Zhipeng and Zhang, Zhongyu and Zhang, Zhen},
   year={2025},
   month=sep, pages={633–638} }

@inproceedings{10.1145/3442381.3450127,
author = {Chen, Jia and Mao, Jiaxin and Liu, Yiqun and Zhang, Fan and Zhang, Min and Ma, Shaoping},
title = {Towards a Better Understanding of Query Reformulation Behavior in Web Search},
year = {2021},
isbn = {9781450383127},
publisher = {Association for Computing Machinery},
address = {New York, NY, USA},
url = {https://doi.org/10.1145/3442381.3450127},
doi = {10.1145/3442381.3450127},
booktitle = {Proceedings of the Web Conference 2021},
pages = {743–755},
numpages = {13},
location = {Ljubljana, Slovenia},
series = {WWW '21}
}

@misc{yao2023reactsynergizingreasoningacting,
      title={ReAct: Synergizing Reasoning and Acting in Language Models}, 
      author={Shunyu Yao and Jeffrey Zhao and Dian Yu and Nan Du and Izhak Shafran and Karthik Narasimhan and Yuan Cao},
      year={2023},
      eprint={2210.03629},
      archivePrefix={arXiv},
      primaryClass={cs.CL},
      url={https://arxiv.org/abs/2210.03629}, 
}

@inproceedings{ma-etal-2025-s2r,
    title = "{S}$^2${R}: Teaching {LLM}s to Self-verify and Self-correct via Reinforcement Learning",
    author = "Ma, Ruotian  and
      Wang, Peisong  and
      Liu, Cheng  and
      Liu, Xingyan  and
      Chen, Jiaqi  and
      Zhang, Bang  and
      Zhou, Xin  and
      Du, Nan  and
      Li, Jia",
    editor = "Che, Wanxiang  and
      Nabende, Joyce  and
      Shutova, Ekaterina  and
      Pilehvar, Mohammad Taher",
    booktitle = "Proceedings of the 63rd Annual Meeting of the Association for Computational Linguistics (Volume 1: Long Papers)",
    month = jul,
    year = "2025",
    address = "Vienna, Austria",
    publisher = "Association for Computational Linguistics",
    url = "https://aclanthology.org/2025.acl-long.1104/",
    doi = "10.18653/v1/2025.acl-long.1104",
    pages = "22632--22654",
    ISBN = "979-8-89176-251-0"
}

@misc{cheng2025agentr1trainingpowerfulllm,
      title={Agent-R1: Training Powerful LLM Agents with End-to-End Reinforcement Learning}, 
      author={Mingyue Cheng and Jie Ouyang and Shuo Yu and Ruiran Yan and Yucong Luo and Zirui Liu and Daoyu Wang and Qi Liu and Enhong Chen},
      year={2025},
      eprint={2511.14460},
      archivePrefix={arXiv},
      primaryClass={cs.CL},
      url={https://arxiv.org/abs/2511.14460}, 
}

@misc{yuan2025agentrtraininglanguagemodel,
      title={Agent-R: Training Language Model Agents to Reflect via Iterative Self-Training}, 
      author={Siyu Yuan and Zehui Chen and Zhiheng Xi and Junjie Ye and Zhengyin Du and Jiecao Chen},
      year={2025},
      eprint={2501.11425},
      archivePrefix={arXiv},
      primaryClass={cs.AI},
      url={https://arxiv.org/abs/2501.11425}, 
}

@misc{yandex2014,
    author = {Eugene and serdyukovpv and Will Cukierski},
    title = {Personalized Web Search Challenge},
    year = {2013},
    howpublished = {\url{https://kaggle.com/competitions/yandex-personalized-web-search-challenge}},
    note = {Kaggle}
}

@inproceedings{10.1145/2910896.2925447,
author = {Mitsui, Matthew and Shah, Chirag},
title = {Coagmento 2.0: A System for Capturing Individual and Group Information Seeking Behavior},
year = {2016},
isbn = {9781450342292},
publisher = {Association for Computing Machinery},
address = {New York, NY, USA},
url = {https://doi.org/10.1145/2910896.2925447},
doi = {10.1145/2910896.2925447},
booktitle = {Proceedings of the 16th ACM/IEEE-CS on Joint Conference on Digital Libraries},
pages = {233–234},
numpages = {2},
location = {Newark, New Jersey, USA},
series = {JCDL '16}
}

@inproceedings{bhattacharya2021yasbil,
author = {Bhattacharya, Nilavra and Gwizdka, Jacek},
title = {YASBIL: Yet Another Search Behaviour (and) Interaction Logger},
year = {2021},
isbn = {9781450380379},
publisher = {Association for Computing Machinery},
address = {New York, NY, USA},
url = {https://doi.org/10.1145/3404835.3462800},
doi = {10.1145/3404835.3462800},
booktitle = {Proceedings of the 44th International ACM SIGIR Conference on Research and Development in Information Retrieval},
pages = {2585–2589},
numpages = {5},
location = {Virtual Event, Canada},
series = {SIGIR '21}
}

@misc{zhan2025evaluatingintelligencetrialerror,
      title={Evaluating Intelligence via Trial and Error}, 
      author={Jingtao Zhan and Jiahao Zhao and Jiayu Li and Yiqun Liu and Bo Zhang and Qingyao Ai and Jiaxin Mao and Hongning Wang and Min Zhang and Shaoping Ma},
      year={2025},
      eprint={2502.18858},
      archivePrefix={arXiv},
      primaryClass={cs.AI},
      url={https://arxiv.org/abs/2502.18858}, 
}

@article{10.3758/s13423-021-02022-8,
    author = {Mera, Yeray and Rodriguez, Gabriel and Marin-Garcia, Eugenia},
    year = {2021},
    month = {11},
    title = {Unraveling the benefits of experiencing errors during learning: Definition, modulating factors, and explanatory theories},
    volume = {29},
    journal = {Psychonomic Bulletin \& Review},
    doi = {10.3758/s13423-021-02022-8}
}

@book{newell2019human,
  title={Human Problem Solving},
  author={Newell, A. and Simon, H.A.},
  isbn={9781635617924},
  url={https://books.google.co.jp/books?id=Gf8EwgEACAAJ},
  year={2019},
  publisher={Echo Point Books and Media}
}

@book{popper1999all,
  title={All Life is Problem Solving},
  author={Popper, Karl},
  year={1999},
  publisher={Routledge},
  address={London},
  isbn={978-0415174862},
  note={Translated by Patrick Camiller}
}

@article{metcalfe2017learning,
  title={Learning from Errors},
  author={Metcalfe, Janet},
  journal={Annual Review of Psychology},
  volume={68},
  number={1},
  pages={465--489},
  year={2017},
  publisher={Annual Reviews},
  doi={10.1146/annurev-psych-010416-044022}
}

@misc{openai2024openaio1card,
      title={OpenAI o1 System Card}, 
      author={OpenAI and : and Aaron Jaech and Adam Kalai and Adam Lerer and Adam Richardson and Ahmed El-Kishky and Aiden Low and Alec Helyar and Aleksander Madry and Alex Beutel and Alex Carney and Alex Iftimie and Alex Karpenko and Alex Tachard Passos and Alexander Neitz and Alexander Prokofiev and Alexander Wei and Allison Tam and Ally Bennett and Ananya Kumar and Andre Saraiva and Andrea Vallone and Andrew Duberstein and Andrew Kondrich and Andrey Mishchenko and Andy Applebaum and Angela Jiang and Ashvin Nair and Barret Zoph and Behrooz Ghorbani and Ben Rossen and Benjamin Sokolowsky and Boaz Barak and Bob McGrew and Borys Minaiev and Botao Hao and Bowen Baker and Brandon Houghton and Brandon McKinzie and Brydon Eastman and Camillo Lugaresi and Cary Bassin and Cary Hudson and Chak Ming Li and Charles de Bourcy and Chelsea Voss and Chen Shen and Chong Zhang and Chris Koch and Chris Orsinger and Christopher Hesse and Claudia Fischer and Clive Chan and Dan Roberts and Daniel Kappler and Daniel Levy and Daniel Selsam and David Dohan and David Farhi and David Mely and David Robinson and Dimitris Tsipras and Doug Li and Dragos Oprica and Eben Freeman and Eddie Zhang and Edmund Wong and Elizabeth Proehl and Enoch Cheung and Eric Mitchell and Eric Wallace and Erik Ritter and Evan Mays and Fan Wang and Felipe Petroski Such and Filippo Raso and Florencia Leoni and Foivos Tsimpourlas and Francis Song and Fred von Lohmann and Freddie Sulit and Geoff Salmon and Giambattista Parascandolo and Gildas Chabot and Grace Zhao and Greg Brockman and Guillaume Leclerc and Hadi Salman and Haiming Bao and Hao Sheng and Hart Andrin and Hessam Bagherinezhad and Hongyu Ren and Hunter Lightman and Hyung Won Chung and Ian Kivlichan and Ian O'Connell and Ian Osband and Ignasi Clavera Gilaberte and Ilge Akkaya and Ilya Kostrikov and Ilya Sutskever and Irina Kofman and Jakub Pachocki and James Lennon and Jason Wei and Jean Harb and Jerry Twore and Jiacheng Feng and Jiahui Yu and Jiayi Weng and Jie Tang and Jieqi Yu and Joaquin Quiñonero Candela and Joe Palermo and Joel Parish and Johannes Heidecke and John Hallman and John Rizzo and Jonathan Gordon and Jonathan Uesato and Jonathan Ward and Joost Huizinga and Julie Wang and Kai Chen and Kai Xiao and Karan Singhal and Karina Nguyen and Karl Cobbe and Katy Shi and Kayla Wood and Kendra Rimbach and Keren Gu-Lemberg and Kevin Liu and Kevin Lu and Kevin Stone and Kevin Yu and Lama Ahmad and Lauren Yang and Leo Liu and Leon Maksin and Leyton Ho and Liam Fedus and Lilian Weng and Linden Li and Lindsay McCallum and Lindsey Held and Lorenz Kuhn and Lukas Kondraciuk and Lukasz Kaiser and Luke Metz and Madelaine Boyd and Maja Trebacz and Manas Joglekar and Mark Chen and Marko Tintor and Mason Meyer and Matt Jones and Matt Kaufer and Max Schwarzer and Meghan Shah and Mehmet Yatbaz and Melody Y. Guan and Mengyuan Xu and Mengyuan Yan and Mia Glaese and Mianna Chen and Michael Lampe and Michael Malek and Michele Wang and Michelle Fradin and Mike McClay and Mikhail Pavlov and Miles Wang and Mingxuan Wang and Mira Murati and Mo Bavarian and Mostafa Rohaninejad and Nat McAleese and Neil Chowdhury and Neil Chowdhury and Nick Ryder and Nikolas Tezak and Noam Brown and Ofir Nachum and Oleg Boiko and Oleg Murk and Olivia Watkins and Patrick Chao and Paul Ashbourne and Pavel Izmailov and Peter Zhokhov and Rachel Dias and Rahul Arora and Randall Lin and Rapha Gontijo Lopes and Raz Gaon and Reah Miyara and Reimar Leike and Renny Hwang and Rhythm Garg and Robin Brown and Roshan James and Rui Shu and Ryan Cheu and Ryan Greene and Saachi Jain and Sam Altman and Sam Toizer and Sam Toyer and Samuel Miserendino and Sandhini Agarwal and Santiago Hernandez and Sasha Baker and Scott McKinney and Scottie Yan and Shengjia Zhao and Shengli Hu and Shibani Santurkar and Shraman Ray Chaudhuri and Shuyuan Zhang and Siyuan Fu and Spencer Papay and Steph Lin and Suchir Balaji and Suvansh Sanjeev and Szymon Sidor and Tal Broda and Aidan Clark and Tao Wang and Taylor Gordon and Ted Sanders and Tejal Patwardhan and Thibault Sottiaux and Thomas Degry and Thomas Dimson and Tianhao Zheng and Timur Garipov and Tom Stasi and Trapit Bansal and Trevor Creech and Troy Peterson and Tyna Eloundou and Valerie Qi and Vineet Kosaraju and Vinnie Monaco and Vitchyr Pong and Vlad Fomenko and Weiyi Zheng and Wenda Zhou and Wes McCabe and Wojciech Zaremba and Yann Dubois and Yinghai Lu and Yining Chen and Young Cha and Yu Bai and Yuchen He and Yuchen Zhang and Yunyun Wang and Zheng Shao and Zhuohan Li},
      year={2024},
      eprint={2412.16720},
      archivePrefix={arXiv},
      primaryClass={cs.AI},
      url={https://arxiv.org/abs/2412.16720}, 
}

@misc{kimiteam2025kimik15scalingreinforcement,
      title={Kimi k1.5: Scaling Reinforcement Learning with LLMs}, 
      author={Kimi Team and Angang Du and Bofei Gao and Bowei Xing and Changjiu Jiang and Cheng Chen and Cheng Li and Chenjun Xiao and Chenzhuang Du and Chonghua Liao and Chuning Tang and Congcong Wang and Dehao Zhang and Enming Yuan and Enzhe Lu and Fengxiang Tang and Flood Sung and Guangda Wei and Guokun Lai and Haiqing Guo and Han Zhu and Hao Ding and Hao Hu and Hao Yang and Hao Zhang and Haotian Yao and Haotian Zhao and Haoyu Lu and Haoze Li and Haozhen Yu and Hongcheng Gao and Huabin Zheng and Huan Yuan and Jia Chen and Jianhang Guo and Jianlin Su and Jianzhou Wang and Jie Zhao and Jin Zhang and Jingyuan Liu and Junjie Yan and Junyan Wu and Lidong Shi and Ling Ye and Longhui Yu and Mengnan Dong and Neo Zhang and Ningchen Ma and Qiwei Pan and Qucheng Gong and Shaowei Liu and Shengling Ma and Shupeng Wei and Sihan Cao and Siying Huang and Tao Jiang and Weihao Gao and Weimin Xiong and Weiran He and Weixiao Huang and Weixin Xu and Wenhao Wu and Wenyang He and Xianghui Wei and Xianqing Jia and Xingzhe Wu and Xinran Xu and Xinxing Zu and Xinyu Zhou and Xuehai Pan and Y. Charles and Yang Li and Yangyang Hu and Yangyang Liu and Yanru Chen and Yejie Wang and Yibo Liu and Yidao Qin and Yifeng Liu and Ying Yang and Yiping Bao and Yulun Du and Yuxin Wu and Yuzhi Wang and Zaida Zhou and Zhaoji Wang and Zhaowei Li and Zhen Zhu and Zheng Zhang and Zhexu Wang and Zhilin Yang and Zhiqi Huang and Zihao Huang and Ziyao Xu and Zonghan Yang and Zongyu Lin},
      year={2025},
      eprint={2501.12599},
      archivePrefix={arXiv},
      primaryClass={cs.AI},
      url={https://arxiv.org/abs/2501.12599}, 
}

@misc{yang2025qwen3technicalreport,
      title={Qwen3 Technical Report}, 
      author={An Yang and Anfeng Li and Baosong Yang and Beichen Zhang and Binyuan Hui and Bo Zheng and Bowen Yu and Chang Gao and Chengen Huang and Chenxu Lv and Chujie Zheng and Dayiheng Liu and Fan Zhou and Fei Huang and Feng Hu and Hao Ge and Haoran Wei and Huan Lin and Jialong Tang and Jian Yang and Jianhong Tu and Jianwei Zhang and Jianxin Yang and Jiaxi Yang and Jing Zhou and Jingren Zhou and Junyang Lin and Kai Dang and Keqin Bao and Kexin Yang and Le Yu and Lianghao Deng and Mei Li and Mingfeng Xue and Mingze Li and Pei Zhang and Peng Wang and Qin Zhu and Rui Men and Ruize Gao and Shixuan Liu and Shuang Luo and Tianhao Li and Tianyi Tang and Wenbiao Yin and Xingzhang Ren and Xinyu Wang and Xinyu Zhang and Xuancheng Ren and Yang Fan and Yang Su and Yichang Zhang and Yinger Zhang and Yu Wan and Yuqiong Liu and Zekun Wang and Zeyu Cui and Zhenru Zhang and Zhipeng Zhou and Zihan Qiu},
      year={2025},
      eprint={2505.09388},
      archivePrefix={arXiv},
      primaryClass={cs.CL},
      url={https://arxiv.org/abs/2505.09388}, 
}

@inproceedings{10.1145/3357384.3358158,
author = {Chen, Jia and Mao, Jiaxin and Liu, Yiqun and Zhang, Min and Ma, Shaoping},
title = {TianGong-ST: A New Dataset with Large-scale Refined Real-world Web Search Sessions},
year = {2019},
isbn = {9781450369763},
publisher = {Association for Computing Machinery},
address = {New York, NY, USA},
url = {https://doi.org/10.1145/3357384.3358158},
doi = {10.1145/3357384.3358158},
booktitle = {Proceedings of the 28th ACM International Conference on Information and Knowledge Management},
pages = {2485–2488},
numpages = {4},
location = {Beijing, China},
series = {CIKM '19}
}

@inproceedings{10.1145/3411764.3445618,
author = {Palani, Srishti and Ding, Zijian and Nguyen, Austin and Chuang, Andrew and MacNeil, Stephen and Dow, Steven P.},
title = {CoNotate: Suggesting Queries Based on Notes Promotes Knowledge Discovery},
year = {2021},
isbn = {9781450380966},
publisher = {Association for Computing Machinery},
address = {New York, NY, USA},
url = {https://doi.org/10.1145/3411764.3445618},
doi = {10.1145/3411764.3445618},
booktitle = {Proceedings of the 2021 CHI Conference on Human Factors in Computing Systems},
articleno = {726},
numpages = {14},
location = {Yokohama, Japan},
series = {CHI '21}
}

@misc{zhang2025qwen3embeddingadvancingtext,
      title={Qwen3 Embedding: Advancing Text Embedding and Reranking Through Foundation Models}, 
      author={Yanzhao Zhang and Mingxin Li and Dingkun Long and Xin Zhang and Huan Lin and Baosong Yang and Pengjun Xie and An Yang and Dayiheng Liu and Junyang Lin and Fei Huang and Jingren Zhou},
      year={2025},
      eprint={2506.05176},
      archivePrefix={arXiv},
      primaryClass={cs.CL},
      url={https://arxiv.org/abs/2506.05176}, 
}

@misc{gou2024criticlargelanguagemodels,
      title={CRITIC: Large Language Models Can Self-Correct with Tool-Interactive Critiquing}, 
      author={Zhibin Gou and Zhihong Shao and Yeyun Gong and Yelong Shen and Yujiu Yang and Nan Duan and Weizhu Chen},
      year={2024},
      eprint={2305.11738},
      archivePrefix={arXiv},
      primaryClass={cs.CL},
      url={https://arxiv.org/abs/2305.11738}, 
}

@misc{ozer2025marmultiagentreflexionimprovesreasoning,
      title={MAR:Multi-Agent Reflexion Improves Reasoning Abilities in LLMs}, 
      author={Onat Ozer and Grace Wu and Yuchen Wang and Daniel Dosti and Honghao Zhang and Vivi De La Rue},
      year={2025},
      eprint={2512.20845},
      archivePrefix={arXiv},
      primaryClass={cs.AI},
      url={https://arxiv.org/abs/2512.20845}, 
}

@inproceedings{Simon1978InformationProcessingTO,
  title={Information-Processing Theory of Human Problem Solving},
  author={Herbert A. Simon},
  year={1978},
  url={https://api.semanticscholar.org/CorpusID:10344827}
}

@book{darwin1859origin,
  title={On the Origin of Species by Means of Natural Selection, Or, The Preservation of Favoured Races in the Struggle for Life},
  author={Darwin, C.},
  lccn={06017473},
  url={https://books.google.co.jp/books?id=jTZbAAAAQAAJ},
  year={1859},
  publisher={J. Murray}
}

@misc{lewis2021retrievalaugmentedgenerationknowledgeintensivenlp,
      title={Retrieval-Augmented Generation for Knowledge-Intensive NLP Tasks}, 
      author={Patrick Lewis and Ethan Perez and Aleksandra Piktus and Fabio Petroni and Vladimir Karpukhin and Naman Goyal and Heinrich Küttler and Mike Lewis and Wen-tau Yih and Tim Rocktäschel and Sebastian Riedel and Douwe Kiela},
      year={2021},
      eprint={2005.11401},
      archivePrefix={arXiv},
      primaryClass={cs.CL},
      url={https://arxiv.org/abs/2005.11401}, 
}
 
\end{document}